\pdfoutput=1
\documentclass[11pt]{article}

\usepackage[final]{acl}

\usepackage{times}
\usepackage{latexsym}
\usepackage{multirow}
\usepackage{amsmath}

\DeclareMathOperator*{\argmax}{arg\,max}
\usepackage{algorithm}
\usepackage{algpseudocode}
\usepackage{natbib}

\usepackage[T1]{fontenc}
\usepackage[utf8]{inputenc}

\usepackage{microtype}

\usepackage{inconsolata}

\usepackage{graphicx}
\usepackage{adjustbox}
\usepackage{booktabs}

\title{Efficient Out-of-Scope Detection in Dialogue Systems via Uncertainty-Driven LLM Routing}

\author{\'Alvaro Zaera\thanks{This work was conducted as part of the author's internship at Telepathy Labs.},  Diana Nicoleta Popa,  Ivan Sekuli\'c, Paolo Rosso \\
Telepathy Labs GmbH, Z\"urich, Switzerland\\
\texttt{\{firstname\}.\{lastname\}@telepathy.ai}}

\begin{document}
\maketitle
\begin{abstract}
Out-of-scope (OOS) intent detection is a critical challenge in task-oriented dialogue systems (TODS), as it ensures robustness to unseen and ambiguous queries. In this work, we propose a novel but simple modular framework that combines uncertainty modeling with fine-tuned large language models (LLMs) for efficient and accurate OOS detection. 
The first step applies uncertainty estimation to the output of an in-scope intent detection classifier, which is currently deployed in a real-world TODS handling tens of thousands of user interactions daily. The second step then leverages an emerging LLM-based approach, where a fine-tuned LLM is triggered to make a final decision on instances with high uncertainty.
Unlike prior approaches, our method effectively balances computational efficiency and performance, combining traditional approaches with LLMs and yielding state-of-the-art results on key OOS detection benchmarks, including real-world OOS data acquired from a deployed TODS. %Experimental results demonstrate that our approach outperforms existing methods %Additional findings support fine-tuning contrary to previous work.
\end{abstract}

%\todo{Add somewhere that contrary to prev work we show that fine-tuning can be done inexpensively.}
\section{Introduction}

Intent detection is a fundamental task in natural language understanding, enabling systems to accurately interpret and respond to user queries by identifying their underlying intention \cite{casanueva-etal-2020-efficient}. While intent detection ensures that in-scope (INS) queries are mapped to predefined intents, detecting out-of-scope (OOS) intents is equally critical, especially in real-world applications, where users often interact in unpredictable ways, by, e.g., posing queries %or statements 
that fall outside the system's designed capabilities~\cite{larson-etal-2019-evaluation,wang2024beyond}. 

Without effective OOS detection, such inputs could lead to incorrect responses, reduced user trust, and eventual system failures as the universe of OOS queries for any TOD system is infinitely large~\cite{arora2024intent}. By identifying OOS queries, systems can gracefully handle such cases, by generating a predefined or dynamic response indicating its inability to process the request, by activating a fallback mechanism such as escalating the conversation to a human agent or by triggering updates to expand system coverage. % This dual focus on in-scope and OOS detection is essential for building robust, reliable, and user-friendly conversational systems in practical settings [cite].

\begin{figure}
    \centering
    \includegraphics[width=1\linewidth]{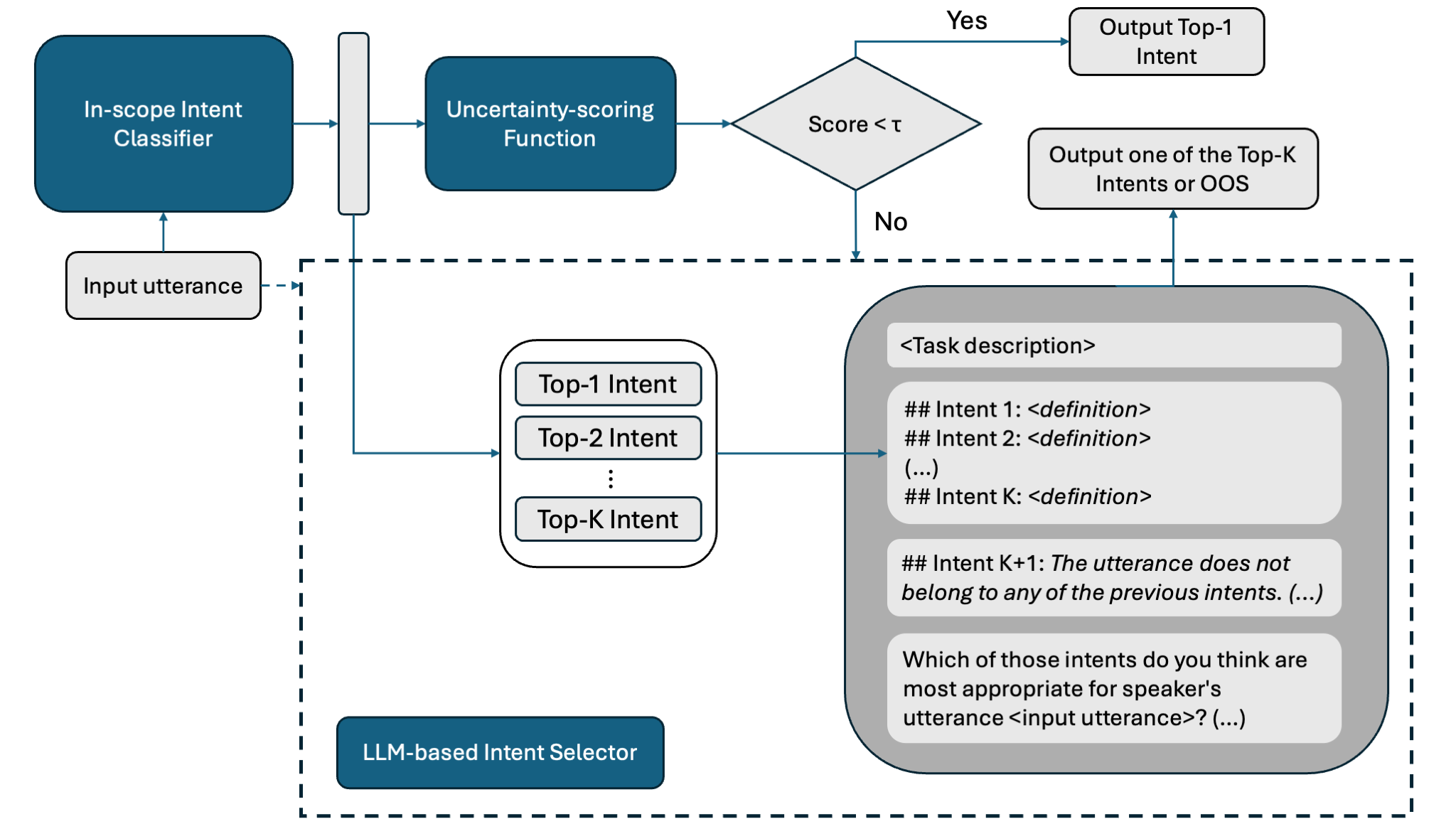}
    \caption{Overview of \textbf{UDRIL}. An \textbf{uncertainty-scoring function} is applied to the output of an \textbf{in-scope classifier}. When a user utterance is potentially out-of-scope, ambiguous or misclassified, as indicated by the uncertainty score and a defined threshold, a \textbf{fine-tuned LLM} is prompted to correct the prediction; otherwise, the classifier's original prediction is maintained.} \label{fig:overall_framework}
\end{figure}

%\todo{introduce it in a more modular way referring to Figure 1 }
To address these challenges, we propose \textbf{Uncertainty-DRIven Large language models triggering, (\texttt{UDRIL})}, a two-step method that combines efficiency with accuracy for robust intent detection. \texttt{UDRIL} is depicted in Figure~\ref{fig:overall_framework} and consists of an in-scope intent classifier, an uncertainty prediction scoring function, and an LLM-based module.
Specifically, we use a BERT-based classifier to ensure both effectiveness and efficiency in a task-oriented dialogue system (TODS) that is currently deployed in production and handling tens of thousands of user interactions daily. To refine predictions, we first apply NNK-Means \cite{gulati-etal-2024-distribution} to identify high-uncertainty instances. For these cases, an emerging LLM-based approach is employed, where a fine-tuned LLM makes the final decision. 
This hierarchical approach leverages the efficiency of the BERT model for the majority of cases, while utilizing the LLM's capabilities for more ambiguous or complex inputs, including OOS detection. Our results demonstrate significant improvements in OOS detection, both on internal real-world dataset and on publicly available data. % (with up to $15\%$ relative improvement compared to previous state of the art). 
Notably, these gains are achieved with additional gains in effectiveness for INS intent detection ($+5\%$), highlighting the method’s overall robustness and practicality.

%Each prediction is then evaluated for uncertainty, and if the model identifies uncertainty, \todo{[rethink ``uncertainty''; write in greater detail what we do (k-means on embeddings)]} the top-k possible intents are passed to a large language model (LLM).

Our main contributions are as follows: %\textbf{(1)} 
\begin{itemize}
    \item a simple modular framework for joint INS and OOS intent detection, combining  strengths of traditional intent classification, uncertainty modeling and LLMs;
    \item a design that selectively escalates user input to a more resource-intensive LLM, balancing efficiency and performance;
    \item state-of-the-art results on publicly available datasets and  on real-world industry data from a deployed system, demonstrating practical applicability and effectiveness.
\end{itemize} %\textbf{(2)}  \textbf{(3)} 
% \todo{add somewhere link to the idea of \cite{arora2024intent}: "intent detection is being used to identify the right knowledge sources, APIs, and tools to call for retrieval augmented generation.". we take a different approach, in which insteda of eliminating the classifier totally, we use bets of both worlds.}

%\textbf{Novel and Simple!}
%We take 3 things: classifier, uncertainty detection, and an LLM, and we make significant improvements!
%Supporting stories:
%\begin{itemize}
%    \item works with limited data --> addresses the cold start problem, lowers the bar for entry to deploy it;
%    \item we can use entropy for uncertainty --> if we don't wanna bother with k-means;
%    \item you can use other guidelines (human-intensive work avoided);
%\end{itemize}

\section{Related Work}
\label{sec:rw}
% \todo{Shorten up RW on intents}
% With the recent advancements in the capabilities of large language models (LLMs), LLM-based intent detection has garnered increasing attention~\citep{parikh2023exploring}. 
% For instance, \citet{hong2024exploring} demonstrate that incorporating natural language intent descriptions enhances the performance of LLMs on zero-shot intent prediction tasks. 
% Building on this, \citet{zhang2024discrimination} report further improvements achieved through fine-tuning LLMs for intent detection.
% Similarly, \citet{mirza2024illuminer} also report increase of a fine-tuned LLM, compared to in-context learning.
% INTRO: INTENT DETECTION IS AN IMPORTANT REAL WORLD TASK
Intent detection is an important task both in TODS~\cite{casanueva-etal-2020-efficient} and in, now emerging, agent-based systems, where we aim to identify the right knowledge sources, APIs, and tools to use~\cite{arora2024intent}.

% THE WORLD BEFORE LLMs (mention post-hoc methods, distance-based, etc.)
\paragraph{Non-LLM-based OOS Intent Detection.} Previous research explored various approaches to intent detection using transformer-based classifiers. A key area of focus has been OOS detection, with methods generally falling into two categories: post-hoc methods that detect OOS instances after obtaining model representations, and approaches that enhance model robustness by modifying the training process %or learning framework 
to better handle OOS data~\cite{gulati-etal-2024-distribution}. We focus on the first category, as these methods are modular, adaptable, and easier to maintain, allowing for easy %<TAG_FINAL>replacement with new methods or 
updates to the architecture without requiring intensive retraining. Particularly relevant in practice is the work by \citet{gulati-etal-2024-distribution}, in which the soft-clustering technique NNK-Means ~\cite{nnkmeans-2022} is applied for OOS detection. This enhances performance while also offering superior computational and memory efficiency compared to previous approaches.

\paragraph{LLM-based Intent Detection.} 
Recently, LLM-based intent detection received significant attention, with studies analyzing the effect in intent detection performance produced by the incorporation of high-quality natural language intent descriptions~\cite{hong2024exploring}. Off-the-shelf LLMs have been shown to outperform non-LLM based methods in few-shot settings where the training set only consist of a small number of utterances per intent class ~\citep{parikh2023exploring}. \citet{hong2024exploring} and \citet{zhang2024discrimination} elaborate on this finding, showing that LLMs fine-tuned on intent detection datasets improve off-the-shelf LLMs, incorporating the ability to detect intents for domains unseen in training. 
% descriptions are beneficial (specially with intents unseen in training) ~\citep{hong2024exploring} ~\citep{parikh2023exploring}
%Fine-tuning has been also proved to be beneficial in few-shot settings (i.e., fine-tuned on the same intents used at inference, until now we have discussed only zero-shot, LLM has not seen those intents in training): 1) allow use smaller models (\citep{parikh2023exploring}) 2) better than ICL ~\citep{mirza2024illuminer}) 3) gives the ability to detect intents for domains unseen in training zhang2024discrimination ~\citep{hong2024exploring} 4) improve near-OOD detection using last-hidden state representation (so it has an implicit impact also for prompting I guess?) \citep{liu2024good} 
Fine-tuning has also proven to be beneficial in few-shot settings, allowing to obtain better results with smaller LLMs compared to off-the-shelf LLMs~\citep{parikh2023exploring} and in-context learning (ICL) approaches~\citep{mirza2024illuminer}). 

%however, this is only in few-shot settings with not too much data. In full data settings LLM are still not clearly better than bert based approaches: ~\citep{mirza2024illuminer}
However, the performance improvement achieved by LLM-based intent detection, as compared to earlier non-LLM methods, is primarily reported in few-shot settings, where the training is strictly constrained by the number of intents per class. Previous studies reporting comparisons in full-data settings show that LLMs still underperform relative to BERT-based approaches in such cases~\citep{parikh2023exploring, mirza2024illuminer}. This underscores the continued relevance of BERT-based methods for practical deployment. Combining the strengths of both LLMs and BERT-based approaches could lead to more flexible systems, capable of adapting to a wider range of training data scenarios and enhancing deployment versatility. 

%Also for OOS detection, LLM are not good by default (point 4. of fine-tuning beneficial show that it helps, but by default it struggles with OOS: \citep{arora2024intent} \citep{wang2024beyond} \citep{liu2024good}
In the context of out-of-scope (OOS) detection, LLMs have been shown to struggle with effective detection when relying solely on text representations without additional training \citep{arora2024intent, wang2024beyond}. To address this limitation, \citet{liu2024good} explore the use of fine-tuning via low-rank adaptation (LoRA)~\citep{DBLP:journals/corr/abs-2106-09685} on INS data, demonstrating that this approach enhances the utility of last-token representations for OOS detection through cosine similarity.

%In conclusion, LLM fine-tuning with good descriptions is the way to go. But the advantages of bert-based methods for settings with more data and for OOS detection shouldn't be dropped. That is why we combine everything (fine-tuning, descriptions and bert classifier with an OOS detection method used as uncertainty-scoring function for routing)
\paragraph{Hybrid Approach.} 
Through the current proposal, we aim to adopt a hybrid approach that combines non-LLM-based OOS intent detection methods with fine-tuned LLMs, leveraging the distinct strengths of the previously discussed methods.
%
% INTRODUCTION OF ARORA (ONLY PAPER TO MENTION THE HYBRID APPROACH BEFORE) AT THE SAME TIME WE INTRODUCE OUR APPROACH (COMBINING STRENGTHS OF TWO PREVIOUS PARAGRAPHS CLASSIFIER+POST-HOC OOS METHODS+LLMS)
% - Mention that we tried to combine both strengths
% - Mention the 3 different settings that Arora evaluates and specify why our method is different
A relevant related work to ours is that of \citet{arora2024intent} who also propose a two-step approach to intent classification, 
%THIS IS NOT TRUE: propose a two-step approach, where a resource-efficient classifier first selects top-k intents, 
albeit involving two LLM passes to determine if an utterance is OOS. Additionally, their proposal requires maintaining a vector storage of last token representations %obtained after applying the LLM to a 
for a set of training examples per intent, performing negative data augmentation and employing multiple runs of monte carlo dropout, making the whole process less scalable.
% compared to simply fine-tuning an LLM on a set of guidelines.%This, added to the fact that a threshold need to be set, makes their method less scalable than fine-tuning on a set of guidelines.
Also, contrary to \citet{arora2024intent} who argue that fine-tuning an LLM for this purpose is impractical and prohibitive from development and maintenance perspective, our experiments as well as related work \cite{hong2024exploring} show that fine-tuning with a set of guidelines is helpful for inference even when the said guidelines are later updated. Therefore, from the maintenance perspective, an update of the intent space and guidelines does not require extra work.

\section{Uncertainty-Driven LLM-based Framework for OOS Intent Detection}
\label{sec:method}

We propose \texttt{UDRIL}, a framework for intent classification and OOS detection, consisting of an in-scope intent classifier and an LLM intent refiner, guided by an uncertainty scoring function $f$. 
The system first employs a classifier to generate an in-scope prediction. If the prediction is deemed confident by $f$, it is used directly; otherwise, the LLM refines it based on the classifier’s output. The proposed framework enhances the cost-efficient classifier by enabling OOS detection while selectively leveraging the LLM, a computationally resource-heavy method, ensuring an accuracy - efficiency balance.

We next describe each component of our framework, noting that they can be replaced based on available resources and performance requirements.

% The system works as follows: initially, the classifier generates an in-scope prediction. If the prediction is considered confident by $f$, it is output directly; otherwise, the process is routed to the LLM, which refines the decision using a prompt derived from the classifier’s output. The main goal of $f$ is to detect utterances that are potentially OOS; however, given the nature of the framework, it also provides an opportunity to refine the predictions of unusual in-scope utterances.

% The classifier is efficient and fast but lacks the ability to detect OOS utterances, while the LLM is more capable but computationally expensive. The proposed framework enhances the classifier by enabling OOS detection in a simple and effective way. Moreover, it maintains efficiency by only using the LLM when necessary, reducing computational costs.

\subsection{In-scope Intent Classifier}
Specifically, given user utterance $u$, the initial classifier's task is to model the probability distribution over a set of $N$ classes $\mathcal{Y}$, %$y \in \mathcal{Y}_N$, %where $\mathcal{Y}=\{y_1,\dots,y_N\}$, 
selecting the one with highest probability as an output: ${\hat{y}}_C = \argmax_{y \in \mathcal{Y}} P_C(y \mid u; \theta_C)$
%\begin{equation}
%    \hat{y}_C = \argmax_{y \in \mathcal{Y}} P_C(y \mid u; \theta_C)
%\end{equation}
\noindent where \( P_C(y \mid u; \theta_C) \) is the classifier's predicted probability distribution and \( \theta_C \) its parameters.
% The initial in-scope classifier is intended to be efficient and scalable, leveraging a simple architecture and training process to meet the demands of real-time applications. While various models can be used, we chose DistilBERT (\citet{DBLP:journals/corr/abs-1910-01108}) due to its strong balance between performance and latency. This makes it well-suited for industry settings where low latency and scalability are essential. However, the framework is flexible and can accommodate other, more complex models if needed for higher accuracy.

In order to meet the demands of low-latency %real-time
applications, we model $P_C$ with DistilBERT~\citep{DBLP:journals/corr/abs-1910-01108}, due to its strong balance between efficiency and effectiveness, making it suitable for an industry setting.
Moreover, the training process only models $\theta_C$ and does not incorporate any methods specific to OOS detection, as this responsibility is entirely managed by the uncertainty-scoring function $f$ and the LLM. Instead, the focus is on training the model to perform general classification tasks efficiently. 
We use focal loss~\citep{ross2017focal} during training to address the intent class imbalance that is likely to occur in the training dataset of real dialogue systems.

\subsection{Uncertainty-Scoring Function}

A function $f$ provides an uncertainty score based on the output of the in-scope classifier, which aims to determine whether the prediction is sufficiently reliable or if further processing by the LLM is required. 
Specifically, score $s_u = f(u)$ indicates the uncertainty score for utterance $u$. 
If $s_u$ exceeds a predefined threshold $\tau$, the utterance is routed to the LLM. Otherwise, the classifier’s prediction is used directly.
%If $s_u$ exceeds a predefined threshold $\tau$, the classifier’s prediction is used directly. Otherwise, the utterance is routed to the LLM.

We model $f$ with EC-NNK-Means~\citep{gulati-etal-2024-distribution}, a soft-clustering based method trained on utterance embeddings to learn a dictionary that minimizes the reconstruction error of the training data. At inference, $s_u$ is the NNK-Means reconstruction error.  In \citet{gulati-etal-2024-distribution}, it is shown that new data with high reconstruction error is more likely to be OOS. We observe that this method also has satisfactory results in identifying potentially misclassified INS data, making it valuable for detecting utterances that require prediction refinement. 
%We model $f$ with a soft-clustering based method, NNK-Means~\citep{gulati-etal-2024-distribution}, which is based on utterance embeddings \todo{@Alvaro add specifics -- embeddings of what are compared to what? Eq?}. \todo{Extracted from the NNK-Means paper \citep{gulati-etal-2024-distribution}: \textit{The dictionary and assignments learned by NNK- Means are optimized to minimize the reconstruc- tion error of the training data. New data that cannot be properly reconstructed using this dictionary, i.e., data with a higher reconstruction error, is more likely to be out-of-distribution. Therefore, we can use the definition of reconstruction error from (5) as an OOD score}. I'm not sure how to include the equation because explaining could take too much space. Btw, I just realized terminology wise that we use EC-NNK-Means (entropy-constrained NNK-Means), which is an improvement that prpose in \citep{gulati-etal-2024-distribution} to not need to select the number of atoms of the dictionary as a hyperparameter, it is automatic. }
In our experiments, we apply EC-NNK-Means to the last output embedding of the DistilBERT \texttt{[CLS]} token. %In addition, we set three different thresholds to simulate possible routing strategies depending on the system requirements.

Threshold $\tau$ can be tuned to route higher, or lower, ratio of utterances to the LLM, balancing the effectiveness and efficiency as needed.
In this work, we experiment with three specific thresholds to showcase its effect on the routing ratio and the overall performance. The selected thresholds define low-routing ($\tau=0.15$), moderate-routing ($\tau=0.10$) and high-routing ($\tau=0.05$) strategies.
\subsection{LLM-Based Intent and OOS Detection}

If the classifier is uncertain, i.e., $s_u > \tau$, the utterance $u$ is forwarded to the LLM to make a final decision. Formally, given the top-$k$ intent candidates \( ( \hat{y}_{(1)}, \dots, \hat{y}_{(k)} ) \), as modeled by $P_C$, the LLM either selects the most appropriate intent among the top-$k$ or determines that $u$ is out-of-scope ($OOS$):
% \begin{equation}
% \begin{split}
%     \hat{y}_{LLM} = \arg\max_{y \in \{ y_1, \dots, y_k, \text{OOS} \}} \\
%     P_{LLM}(y \mid  x, y_1, \dots, y_k; \theta_{LLM})
% \end{split}
% \end{equation}
\begin{equation}
\begin{split}
    \hat{y}_{LLM} = \argmax_{y \in \{ \hat{y}_{(1)}, \dots, \hat{y}_{(k)}, OOS \}} \\
    P_{LLM}(y \mid  u, \hat{y}_{(1)}, \dots, \hat{y}_{(k)}; \theta_{LLM})
\end{split}
\label{eq:eq_llm}
\end{equation}

% When deemed necessary, the framework uses an LLM to make the final decision in the intent classification process. 
% The LLM is prompted to select the most appropriate option from the top $k$ intent classes provided by the initial classifier. In addition to these $k$ options, an extra one is included to capture cases where the input does not align with any of the predefined choices. If the LLM selects this extra option, the utterance is considered to be OOS. Thus, the framework design requires the intent classifier to have a good enough top-$k$ performance.

% \todo{Should we include the exact prompt? Should we include the exact prompt of the description generation method?}

In this work, we learn $\theta_{LLM}$ of $P_{LLM}$ via fine-tuning using LoRA~\citep{DBLP:journals/corr/abs-2106-09685} with a language modeling objective. %Algorithm \ref{alg:fine-tuning} presents the creation of the dataset used for fine-tuning the LLM. 
Our method is designed to provide the LLM with OOS detection capabilities using only INS data. For the dataset creation, given each <utterance-gold label> pair $(u,y_u)$, we additionally create one negative example $(u,OOS)$, using $k$ candidates $(y'_{(1)},\dots,y'_{(k)})$ sampled from $\mathcal{Y} \setminus \{y_u\}$, as described in Algorithm \ref{alg:fine-tuning}. %The model is trained to use the OOS label as a way to indicate that the utterance $u$, having a ground truth label $y_u$ does not belong to any of the $k$ presented classes, $y_{(1)},\dots,y_{(k)}$.
We then train using the obtained dataset $D'$ to maximize Eq. (\ref{eq:eq_llm}).

\begin{algorithm}
\caption{Fine-tuning Dataset Creation}
\label{alg:fine-tuning}
\begin{algorithmic}
\State \textbf{Input:} INS Dataset $D$, Classifier $P_C$, Param $\theta_{C}$% , Trained Classifier Parameters $\theta_{C}$ %, LLM $P_{LLM}$
\State \textbf{Output:} Fine-tuning Dataset $D'$
%\State \textbf{Output:} LLM Parameters $\theta_{LLM}$
\State %\textbf{Create training dataset from} $D_{\text{INS}}$
\State \textbf{Initialize:} $D' \gets \emptyset$
\For{each $(u, y_u)$ in $D$}
    \State Use $P_C(\;\cdot\;|u;\theta_{C})$ to obtain \( ( \hat{y}_{(1)}, \dots, \hat{y}_{(k)} ) \)
    \State Add \( (u, (\hat{y}_{(1)}, \dots, \hat{y}_{(k)} ), y_u) \) to $D'$
    \State Sample $k$ distinct intents from $\mathcal{Y} \setminus \{y_u\}$: \\ \hspace{1cm} \( ( y'_{(1)}, \dots, y'_{(k)} ) \) 
    \State Add \( (u, (y'_{(1)}, \dots, y'_{(k)} ), OOS) \) to $D'$
\EndFor
\State \textbf{Return:} Fine-tuning Dataset $D'$
% \State Train using \(D_{\text{OOS}} = \{(u, (y_{(1)}, \dots, y_{(k)} ), y_u)\}\)
% \For{every epoch}
%     \State Shuffle the order of \( ( y_{(1)}, \dots, y_{(k)} ) \) in $D_{\text{OOS}}$
%     \State Update $\theta_{LLM}$ to maximize \\ \hspace{1cm} \( P_{LLM}(y_u \mid  u, y_{(1)}, \dots, y_{(k)}; \theta_{LLM})\)
% \EndFor
%\State Train using \(D_{\text{OOS}}\). We refer to its elements as \(\{(u, (z_{(1)}, \dots, z_{(k)} ), z_u)\}\)
%\For{every epoch}
%    \State Shuffle the order of \( ( z_{(1)}, \dots, z_{(k)} ) \) in $D_{\text{OOS}}$
%    \State Update $\theta_{LLM}$ to maximize \\ \hspace{1cm} \( P_{LLM}(z_u \mid  u, z_{(1)}, \dots, z_{(k)}; \theta_{LLM})\)
%\EndFor
%\State \textbf{Return:} LLM Trained Parameters $\theta_{LLM}$
\end{algorithmic}
\end{algorithm}

% To enable OOS detection each example is presented twice in every epoch of the fine-tuning process. 
% In one presentation, the top-$k$ intents from the classifier are included in the prompt.
% \todo{The ground truth (GT) intent is present as part of these options since the classifier has been trained with these utterances.} This allows the LLM to understand which utterances belong to each intent while also being exposed to plausible confusions by presenting the other top options from the classifier. In the second presentation, $k$ random non-GT intent classes are used, enabling the LLM to learn when an input does not belong to any of the predefined classes. This mechanism implicitly provides the framework with OOS detection capabilities. The order of the $k$ intent classes used in the prompt is randomized at each epoch.

For our experiments, we use Llama 3.1-8B~\citep{dubey2024llama} as the LLM with $k=3$ intent descriptions. The prompt contains a description of each of the $k$ intents. Each epoch, the order of the $k$ candidates is shuffled in the prompt. The fine-tuning set is created using 5 random utterances from the training set per intent class. In cases where the number of available utterances was lower than 5, we performed data augmentation. Having a limited number of examples, combined with using a parameter-efficient fine-tuning technique (LoRA), facilitates deployment in production environments. 

% Following related work, the $k$ intent classes are presented in the prompt along with a description of each option. These descriptions are generated using \todo{ChatGPT given a list of examples per intent class} \todo{What about our data descriptions?} \todo{Mention something about the importance of the quality found in previous work?}. The intent classes are presented in an anonymized way to avoid any bias that a specific intent name might introduce. The options are always displayed as ``Intent 1'', ``Intent 2'', ..., ``Intent $k+1$''.
 
\subsection{Evaluation Setup and Data}
\paragraph{Internal benchmark.} 
Our main goal is to tackle intent detection in our deployed TOD system; thus, we primarily evaluate our approach on an internal benchmark.
To this end, we extract $6492$ real user utterances from our past user-system interactions and manually annotate them with one of $42$ intents.
We refer to this dataset as \emph{BookData}.

\paragraph{Public benchmark.} To ensure comparability to related work, we further evaluate our methods on the real-world data from the HINT3 collection~\citep{arora-etal-2020-hint3}, created from live chatbot interactions in diverse domains. The collection contains three datasets: \emph{SOFMattress} (mattress products retail), \emph{Curekart} (fitness supplements retail), and \emph{Powerplay11} (online gaming).
Utterances in the train sets are labeled with between 21--57 INS intents, while the test sets additionally contain a large number of OOS utterances.
%<TAG_FINAL>We use F1 score as the main evaluation metric to evaluate the classification performance, both INS and OOS. 

\paragraph{Intent guidelines.}
While for internal data, we have access to annotation guidelines, for public benchmarks such guidelines are not made available. To solve this, we generate guidelines for each of the public datasets using OpenAI's GPT3.5: for each intent, we provide as input the intent name and all utterances that are part of the train set for that intent. 
We then ask the LLM to generate a definition such that, when presented along with such examples, a human would choose to label the examples with the given intent. We make no further adjustments or post-processing to the obtained guidelines.
%<TAG_FINAL>We then ask the LLM to generate annotation guidelines such that, when presented with such examples and the generated guidelines, a human would choose to label the examples based on the guidelines with the given intent. We make no further adjustments or post-processing to the obtained guidelines.
    
%Similar to \citet{arora2024intent}, we use  datasets from ) due to their closeness to the conditions of a real dialogue system. In addition, we use an internal dataset. \not-todo{Mention the domain of the intents, number of intents, challenges of every dataset}

% -  Curekart 28 intents fitness supplements retail Train: 599 INS Test: 459 INS and 532 OOS
% - Sofmattress 21 intents mattress products retail Train: 328 INS Test: 253 INS 144 OOS
% - Powerplay11 57 intents online gaming Train: 471 INS Test: 309 INS 674 OOS
% - BookData 42 intents automotive domain Train: 165280 INS 8357 OOS Test: 6045 INS 447 OOS

% To facilitate a more detailed analysis of OOS detection, we also present the F1 score separately for INS and OOS categories, and the OOS precision and recall. \todo{is this true?}

\section{Results and Discussion}
\label{sec:results}

\begin{table*}[]
\centering
\begin{adjustbox}{width=1\textwidth}
\begin{tabular}{@{}lcccc|c|l@{}}
\toprule
Method  & Curekart & SOFMattress & PowerPlay11 & Avg Score & BookData & Param\\ \midrule \midrule
%Large LLMs \\ \midrule
SNA~\cite{arora2024intent} & 0.709 &  0.672  &  0.639 & 0.673 & - & NA\\
Mistral-7B~\cite{arora2024intent}  &  0.615   &  0.699  & 0.384  & 0.566 & - & 7B\\
Claude v3 Haiku~\cite{arora2024intent}    & 0.775   & \textbf{0.815}   & 0.646  & 0.745 & - & NA\\ 
Mistral Large~\cite{arora2024intent}    & 0.779   & 0.767   &  0.668 & 0.738 & - & 123B\\ \midrule
SNA + Claude v3 Haiku~\cite{arora2024intent}    & 0.756   & 0.730   & 0.690  & 0.725 & - & NA\\ 
SNA + Mistral Large~\cite{arora2024intent}   &  0.761  & 0.719   & 0.692  & 0.724  & - & NA\\ 
\midrule
Mistral-7B-2steps~\cite{arora2024intent}  &  0.766   &  0.751  & \textbf{0.739}  & 0.752 & - & 7B\\ \midrule \midrule
%SetFit ~\cite{arora2024intent}      & 0.511   & 0.632   & 0.612  &  0.585       \\ \midrule
%DistilBERT (INS)    &    &    &   & \\
%DistilBERT (All)    &    &    &   &  \\
%\begin{tabular}[c]{@{}l@{}}Llama 3.1-8B\\ (no fine-tuning) (Ours)\end{tabular} &    0.65      &    0.67        &   0.55     &  0.623   \\
%\begin{tabular}[c]{@{}l@{}}Llama 3.1-8B \\ (fine-tuned) (Ours)\end{tabular}    &   0.787       &    0.777    &     0.708   & 0.757 \\ \midrule
%\multicolumn{5}{l}{\textit{Our proposals}} \\ \midrule
UDRIL-noFT (low-route) & 0.637 & 0.661 & 0.525 & 0.607 & 0.831 & 8B\\
UDRIL-noFT (moderate-route) & 0.660 & 0.672 & 0.542 & 0.624 & 0.826 & 8B\\
UDRIL-noFT (high-route) & 0.662 & 0.676 & 0.547 & 0.628 & 0.790 & 8B\\ 
UDRIL-noFT (full-route) &    0.655      &    0.669        &   0.545     &  0.623 & 0.748 & 8B \\ \midrule
UDRIL-FT (low-route)  & 0.727     &  0.764        &     0.677         &     0.722    &  0.852 &  8B  \\
UDRIL-FT (moderate-route)& 0.779   &   0.777       &    0.701          &   0.752   &  \textbf{0.857} &   8B     \\
UDRIL-FT (high-route) & \textbf{0.791}   &   \underline{0.784}       &   \underline{0.710}          &   \textbf{0.761}     &  \underline{0.853}  &  8B    \\ 
UDRIL-FT (full-route) &   \underline{0.787}       &    0.777    &     0.708   & \underline{0.757} &  0.850 & 8B\\ \bottomrule
\end{tabular}
\end{adjustbox}
\caption{F1 scores across state-of-the-art methods and our proposed solution \texttt{UDRIL}, with different routing strategies. The postfix \textit{-noFT} refers to off-the-shelf models that were not fine-tuned, while \textit{-FT} refers to the fine-tuned version of Llama 3.1-8B. \textit{Mistral-7B} is the model proposed in \citet{arora2024intent}, with comparable number of parameters to our method, while \textit{Claude v3 Haiku} and \textit{Mistral Large} are the best performing models of \citet{arora2024intent} - albeit much bigger than our proposed solutions. \textit{SNA + Mistral Large}; and \textit{SNA + Claude v3 Haiku} are hybrid models and \textit{Mistral-7B-2steps} is the best OOS model in \cite{arora2024intent}. Best scores are in \textbf{bold}, second best are \underline{underlined}.}
\label{tab:main_results}
\end{table*}

Table~\ref{tab:main_results} presents results on HINT3 public datasets, comparing state-of-the-art solutions~\cite{arora2024intent} and our methods. 
%We report F1-scores, on average, across all datasets, and per dataset. Additionally, Table \ref{tab:results_ins} showcases our performance on INS data, while Table \ref{tab:results_oos_recall} focuses on OOS performance: to enable comparison to related work, we follow the approach of \cite{arora2024intent} and report OOS recall at best F1-score.
We compare to three main categories of related work results: (1) non-LLM (\emph{SNA}) and the best performing LLM-based approaches in \citet{arora2024intent}: \emph{Mistral-7B}, \emph{Claude v3 Haiku} and \emph{Mistral Large}; (2) hybrid models and (3) the proposal of \citet{arora2024intent} specifically designed for OOS intent detection. 

% \todo{The amazon paper says: \textit{If needed, threshold in Step 2 of our methodology can be chosen such that drop in in-scope performance is less than an upper limit which in-turn would limit the gains in OOS performance though.} Our fine-tuning method works in a way that OOS detection does not depend that strongly on setting a threshold that would reduce INS performance (this was true with INS f1-score; however with accuracy there is a bigger drop if we rout more). In their final OOS method they only beat us because the OOS recall is 0.95 in a dataset where 68\% are OOS.}

\subsection{Open-Source Data}

Average F1-scores across all datasets show that \texttt{UDRIL} provides an average of 2-3\% relative improvement compared to state-of-the-art methods that employ significantly larger LLMs, up to 13\% relative improvement compared to traditional classifier-based approaches and up to 34\% relative improvement when compared to similar-sized LLMs (see comparison to \emph{Mistral-7B} \cite{arora2024intent}). The increase in performance holds regardless of the routing strategy employed. It also holds when using an LLM that was not fine-tuned for the task compared to other similar-sized LLMs (\texttt{UDRIL-noFT} can yield up to 10\% increase compared to \emph{Mistral-7B} \cite{arora2024intent}), validating the value of %<TAG_FINAL> the proposed architecture beyond fine-tuning. 
our architecture beyond fine-tuning.

\texttt{UDRIL} also outperforms hybrid approaches by up to 5\%, despite these latter ones using much larger LLMs. Methodology-wise, \texttt{UDRIL} is also simpler: there is no need for negative data augmentation for the classifier or multiple uncertainty estimation runs, unlike other hybrid proposals.

Finally, \texttt{UDRIL} yields improvements over the OOS-specific method of \citet{arora2024intent} for Curekart and SOFMattress and incurs only slight degradation in the case of PowerPlay11, making it on average the better performing model of the two. Beyond performance, the simplicity of \texttt{UDRIL} also makes it easier to use in practice.

\subsection{Real-World Data}
We observe a performance increase on \emph{BookData} when fine-tuning is employed and a progressive decrease as we route more utterances with the non-fine-tuned models.
These results suggest that, for a real-world industry setting, fine-tuning LLM-based models on in-domain labeled data is still superior to switching to in-context learning with LLMs. %Since fine-tuning was done using only a limited set of 5 utterances per label, this also means it's production-friendly.

Furthermore, increasing the amount of training data, even with noisy labels, improves the performance of a DistilBERT-based classifier, thereby reducing the need for extensive routing to achieve optimal results. Additionally, fine-tuning the LLM on a small set of utterances enhances the framework's robustness across various routing strategies, enabling effective out-of-scope (OOS) handling without compromising in-scope (INS) performance.

\begin{table*}[]
\centering
\begin{adjustbox}{width=0.9\textwidth}
\begin{tabular}{llcccc}
\toprule
                                                                        % &                         & \multicolumn{1}{l}{Overall Accuracy} & \multicolumn{1}{l}{F1 Score} & \multicolumn{1}{l}{INS Accuracy} & \multicolumn{1}{l}{OOS Recall} \\ \midrule \midrule
                                                                        &                         & \multicolumn{1}{l}{F1 Score} & \multicolumn{1}{l}{INS Accuracy} & \multicolumn{1}{l}{OOS Precision} & \multicolumn{1}{l}{OOS Recall} \\ \midrule \midrule
\multirow{2}{*}{\begin{tabular}[c]{@{}l@{}}SOF\\ Mattress\end{tabular}} %& Mistral-7B (one step)   & 0.705            & 0.699    & \textbf{0.842}            & 0.465               \\
                                                                        % & Mistral-7B-2steps \cite{arora2024intent}  & 0.748            & 0.751    & \textbf{0.767}            & 0.715               \\ %\cline{2-6} 
                                                                        & Mistral-7B-2steps~\cite{arora2024intent}  & 0.751            & \textbf{0.767}    & -            & 0.715               \\ %\cline{2-6} 
                                                                        %& UDRIL-noFT (high-route) & 0.690            & 0.676    & 0.818            & 0.465               \\
                                                                        % & UDRIL-FT (high-route)   & \textbf{0.788}            & \textbf{0.784}    & 0.759            & \textbf{0.840}               \\ \midrule %\midrule
                                                                        & UDRIL-FT (high-route)   & \textbf{0.784}            & 0.759    & 0.725            & \textbf{0.840}               \\ \midrule %\midrule
\multirow{2}{*}{Curekart}                                               %& Mistral-7B (one step)   & 0.601            & 0.615    & \textbf{0.863}           & 0.376               \\
                                                                        % & Mistral-7B-2steps \cite{arora2024intent} & 0.761            & 0.766    & 0.736            & \textbf{0.782}               \\ %\cline{2-6}
                                                                        & Mistral-7B-2steps~\cite{arora2024intent} & 0.766            & 0.736    & -         & \textbf{0.782}               \\ %\cline{2-6}
                                                                        %& UDRIL-noFT (high-route) & 0.650            & 0.662    & 0.854            & 0.474               \\
                                                                        % & UDRIL-FT (high-route)   & \textbf{0.784}            & \textbf{0.791}    & \textbf{0.830}            & 0.744               \\ \midrule %\midrule
                                                                        & UDRIL-FT (high-route)   & \textbf{0.791}            & \textbf{0.830}    &  0.888          & 0.744               \\ \midrule %\midrule
\multirow{2}{*}{\begin{tabular}[c]{@{}l@{}}Power\\ Play11\end{tabular}} %& Mistral-7B (one step)   & 0.357            & 0.384    & \textbf{0.689}            & 0.205               \\
                                                                        % & Mistral-7B-2steps \cite{arora2024intent} & \textbf{0.780}           & \textbf{0.739}    & 0.411            & \textbf{0.950}               \\ %\cline{2-6} 
                                                                        & Mistral-7B-2steps~\cite{arora2024intent} & \textbf{0.739}           & 0.411    & -           & \textbf{0.950}               \\ %\cline{2-6} 
                                                                        %& UDRIL-noFT (high-route) & 0.486            & 0.547    & 0.605            & 0.432               \\
                                                                        % & UDRIL-FT (high-route)   & 0.688            & 0.710    & \textbf{0.557}            & 0.748               \\ \hline
                                                                        & UDRIL-FT (high-route)   & 0.710           & \textbf{0.557}    &    0.857       & 0.748               \\ \hline
                                                                    %\multirow{2}{*}{\begin{tabular}[c]{@{}l@{}}BookData\end{tabular}}
                                                                        %& UDRIL-noFT (moderate-route)   & 0.826           & 0.857    &    0.591       & 0.398 \\
                                                                        %& UDRIL-FT (moderate-route)   & 0.857           & 0.865    &    0.531       & 0.682               \\ \hline
\end{tabular}
\end{adjustbox}
\caption{Best-performing \citet{arora2024intent} method vs UDRIL, focusing on OOS and INS performance.}
\label{tab:oos_results}
\end{table*}

\subsection{Impact of Fine-Tuning on Performance}
Fine-tuning improves the OOS detection capabilities of \texttt{UDRIL} by substantially increasing recall, with only a minor reduction in precision. For instance, in BookData with the full-route setting, the OOS recall increases from 0.403 to 0.698 and the precision is very similar, dropping from 0.514 to 0.508.
%\todo{The reduction in OOS precision can translate into a slight reduction in INS Accuracy (happens for somattress and powerplay, not for curekart and bookdata) (However the F1 INS score remains similar for curekart and powerplay and better for the others) -> We incorporate OOS detection capabilities without a significant INS performance degradation (compute numbers)}
The reduction in OOS precision could potentially lead to a slight decrease in INS performance. This is not the case for BookData, where the INS Accuracy increases from 0.768 with the off-the-shelf LLM to 0.856 with the fine-tuned version in the full-route setting. However, in the HINT3 dataset we do observe slight drops: Curekart 0.817 to 0.815, Sofmattress 0.806 to 0.743 and Powerplay11 0.599 to 0.547. 
%\todo{<TAG_FINAL>: remove this line: it's true that is saying good thongs of our method but it feels more like an unnecessary justification of our method being valid and ends up being more confusing} Nonetheless, this performance drop is not observed when analyzing the F1-score over the subset of INS utterances. In fact, Powerplay11 is the only one where the INS F1-score drop is observed, and it is negligible (from 0.618 to 0.612). 
%
%<TAG_FINAL> A correlation is observed: cases where the first-stage classifier performs worse (such as Powerplay11) are more likely to negatively impact INS performance when incorporating the OOS detection capabilities through fine-tuning.
%A correlation is 
We observe that incorporating OOS detection capabilities through fine-tuning is more likely to negatively impact INS performance for
cases where the first-stage classifier performs worse (such as Powerplay11). %It is important to note also that from a cost perspective,
\subsection{Balancing INS and OOS performance}
Table \ref{tab:oos_results}  compares \texttt{UDRIL} with the method specifically designed for OOS detection in \citet{arora2024intent}. Our approach strikes a better balance between OOS recall and INS accuracy, leading to a superior overall F1 score on two out of three datasets. Powerplay11 is the only exception, where \citet{arora2024intent} outperforms ours. However, this can be attributed to the fact that $\sim$68\% of the utterances in the test split of Powerplay11 are OOS. Their method, which achieves a significantly high OOS recall at the cost of excessively low INS accuracy, has limited practical applicability compared to our more balanced approach. That said, our approach does not achieve ideal INS accuracy either - most likely due to the first-stage classifier: %<TAG_FINAL>This is due to the fact that, as previously mentioned, the performance of our framework is constrained by the first-stage classifier.
since Powerplay11’s training set is of lower quality, this directly impacts both the DistilBERT classifier and the overall performance of the framework.

%Since \texttt{UDRIL} is modular and either of the base classifier, uncertainty measure and LLM choice can be changed, due to space limitations, we only briefly report here some findings.

\textbf{Intent guidelines} Experiments showed that fine-tuning using guidelines of one dataset can be beneficial across datasets: 
%<TAG_FINAL results on SOFMattress and PowerPlay11 when \texttt{UDRIL} is fine-tuned with guidelines suited for the Curekart dataset are comparable to results obtained when the fine-tuning is done using their guidelines directly.
results on SOFMattress and PowerPlay11 with \texttt{UDRIL} fine-tuned using Curekart-specific guidelines are comparable to those obtained when fine-tuning using their own guidelines directly.
These findings are in line with recent work \cite{hong2024exploring} and support the usability of the method in the lack of up-to-date dataset-specific guidelines at fine-tuning time. %\todo{Add that we refer 'for fine-tuning', dataset-specific guidelines are always required at some point. Mention that also allows to update or modify guidelines without requiring to fine-tune again? This is strong because it counterarguments the only reason why they don't fine-tune in the Amazon one.}. 
%\todo{<TAG_FINAL removing this sentence since we already comment on that in the related work section} However, this contradicts the arguments of \cite{arora2024intent}, showing that fine-tuning the LLM can be done inexpensively: even in the case of guidelines mismatch, one can still benefit from fine-tuning on an older version of the guidelines. 

\textbf{Uncertainty measures and LLMs.} We experimented with different LLMs, including recent \textit{DeepSeek-R1-Distill-Llama-8B} and \textit{DeepSeek-R1-Distill-Qwen-7B} models\footnote{https://huggingface.co/deepseek-ai}, as well as several uncertainty measures, such as Shannon Entropy and Energy, as proposed in \citep{SUN2024111167}. Results were similar to the reported ones with some degradation observed when using other uncertainty measures. %\todo{similar comment that I had at the very beginning. These "uncertainty measures" are just non-LLM-based OOS detection metods. I feel we need to make that clear at some point, not sure where. Maybe in 3.3?}

\textbf{How good is our routing strategy?} 
We observe routing strategies above \textit{moderate} yield improvements over existing models, with the preferred approach consisting in \textit{high} amount of routing. 
%<TAG_FINAL> Unsurprisingly, routing all utterances can slightly degrade the performance as originally good classifier predictions may end up being altered by the LLM.

\begin{figure}
    \centering
    \includegraphics[width=\linewidth]{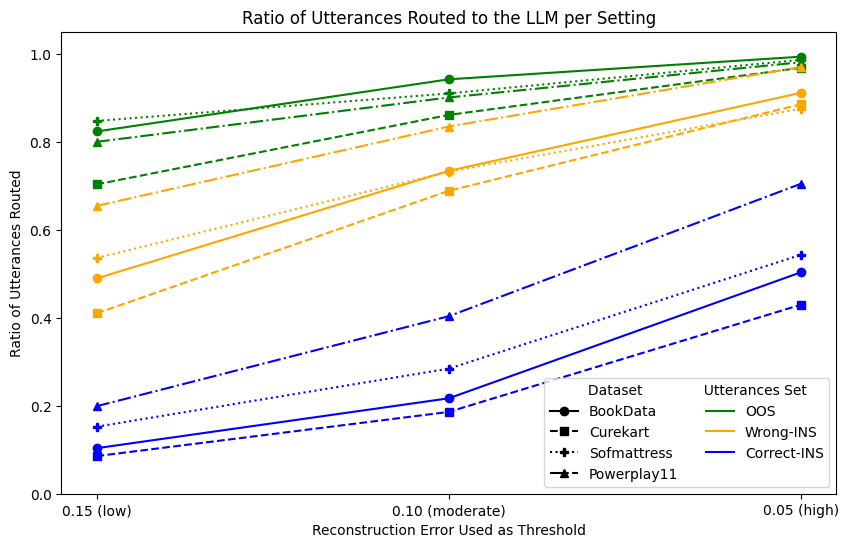}
    \caption{Impact of routing threshold to number of routed utterances across four datasets and three utterance label sets.}
    \label{fig:routing}
\end{figure}

%<TAG_FINAL>Investigating the amount of data that gets routed by our method, we observe that 
The percentage of routed OOS utterances varies between 70-96\% for Curekart, 84-98\% for SOFMattress and 79-98\% for PowerPlay11, depending on how conservative we are. Furthermore,  of the incorrectly labeled INS utterances, our method routes between 40-88\% in the case of Curekart, 53-87\% for SOFMattress and 65-97\% for Powerplay11, as seen from Figure~\ref{fig:routing}. We also observe that when DistilBERT performs better, fewer correctly classified INS utterances are routed to the LLM, demonstrating that the routing method effectively captures prediction uncertainty.
We conclude that our routing method benefits both OOS and INS labels.

\section{Conclusion}
\label{sec:conclusions}

In this paper, we introduce \texttt{UDRIL}, a framework that achieves state-of-the-art performance in both in-scope (INS) intent classification and out-of-scope (OOS) intent detection. Unlike approaches that require modifying or retraining the base intent classifier, \texttt{UDRIL} operates by modeling its outputs, enabling OOS detection while preserving the efficiency of the existing classifier. This makes our framework particularly well-suited for real-world deployment, as shown by the results on our internal benchmark, derived from real user-system interactions, where maintaining low latency and computational efficiency is crucial.

Moreover, \texttt{UDRIL} is modular, allowing for the seamless substitution of different components: base classifier, uncertainty estimation method, and LLM. Furthermore, it provides a practical mechanism for controlling efficiency-performance trade-offs by adjusting the routing percentage threshold, ensuring adaptability to varying production constraints.
By enabling reliable OOS detection without disrupting existing intent classification models, our approach offers a scalable solution for enhancing the robustness of deployed TOD systems.
%It also obtains better results than methods employing similar-sized LLMs, regardless of fine-tuning and / or routing strategy employed. Our proposal outperforms existing models in literature that employ much bigger LLMs on the HINT3~\cite{arora-etal-2020-hint3} public datasets. Finally, we

%<TAG_FINAL>\section*{Limitations and future work}
%One possible limitation of our approach is the fixed selection of top-$k$ intents (where $k=3$) from the classifier routed to the LLM. While initial experiments suggested this choice was reasonable, in the future we plan to include a more systematic evaluation into the optimal setting for different scenarios.

%Another potential limitation of our approach is its reliance on the quality of the intent descriptions used in the prompt. We do not provide at this point an analysis of how the quality of these descriptions and the design of intent classes impact the system's performance, but we plan to look into these aspects in future work.

%Another potential limitation of our approach is the lack of threshold tuning for routing decisions. This was primarily due to the absence of sufficiently valid training and development sets for systematic optimization in HINT3. In future work, obtaining more reliable evaluation data, similar to the one we have internally, could enable fine-tuning these thresholds to improve overall performance.

\section*{Ethical Considerations}
We prioritize user privacy and ensure that no real conversations are reported in this paper. Additionally, we do not release any data or model weights trained on user interactions. All data used in our study was collected with user consent, ensuring ethical use and compliance with the US privacy considerations.

%\section*{Acknowledgments}
%Acknowledgments.

\bibliography{acl_latex}

%\appendix

%\section{Example Appendix}
%\label{sec:appendix}

%This is an appendix.

\end{document}